# Linear Parameter Varying Model Identification for Control of Rotorcraft-based UAV

**Agus Budiyono†**

Department of Aeronautics and Astronautics
Institut Teknologi Bandung
Bandung, Indonesia
†agus.budiyono@ae.itb.ac.id

**H.Y Sutarto**

Research and Development Center
Global Asia Teknologi
Bandung, Indonesia

**ABSTRACT**

A rotorcraft-based unmanned aerial vehicle exhibits more complex properties compared to its full-size counterparts due to its increased sensitivity to control inputs and disturbances and higher bandwidth of its dynamics. As an aerial vehicle with vertical take-off and landing capability, the helicopter specifically poses a difficult problem of transition between forward flight and unstable hover and vice versa. The most traditional approach for solving this nonlinear problem is by linearizing the vehicle dynamics over a set of operating points where a gain scheduling mechanism is then applied. The drawback of this approach is that the control design based on the linearized dynamics might not perform well or worse guarantee stability when it is applied beyond the vicinity of equilibrium. In contrast to this approach, the LPV control technique explicitly takes into account the change in performance due to the real-time parameter variations. The technique therefore theoretically guarantees the performance and robustness over the entire operating envelope. In one of many robust control approaches, one can conduct numerical linearizations of the dynamics of the nonlinear helicopter model made for varying forward speed. This set of linear models is then used to construct an LFT representation of the system by element-wise interpolation of the stability and control derivatives matrices. Despite its guaranteed performance, this approach however is unnecessarily conservative since the LFT does not consider the rate of parameter variation. In this study, we investigate a new approach implementing model identification for use in the LPV control framework. The identification scheme employs recursive least square technique implemented on the LPV system represented by dynamics of helicopter during a transition. The airspeed as the scheduling of parameter trajectory is not assumed to vary slowly. The exclusion of slow parameter change requirement allows for the application of the algorithm for aggressive maneuvering capability without the need of expensive computation. The technique is tested numerically and will be validated in the autonomous flight of a small scale helicopter.

## INTRODUCTION

The stabilization and control for a miniature helicopter has been designed using different techniques. In the1990s, the classical control systems such as single-input-single-output SISO proportional-derivative (PD) feedback control systems have been primarily used. Their controller parameters were typically tuned empirically. This trial-and-error approach to design an "acceptable" control system however is not agreeable with complex multi-input multi-output MIMO systems with sophisticated performance criteria. For more advanced multivariable controller synthesis approaches, an accurate model of the dynamics is required. Such models, however, are not usually readily available and are difficult to develop [1]. To control a model helicopter as a complex MIMO system, an approach that can synthesize a control algorithm to make the helicopter meet performance criteria while satisfying some physical constraints is required. More recent development in this area include the use of optimal control (Linear Quadratic Regulator) implemented on a small aerobatic helicopter designed at MIT [2-3]. Similar approach based on µ-synthesis has been also independently developed for a rotor unmanned aerial vehicle at UC Berkeley [4]. An adaptive high-bandwidth helicopter controller algorithm was synthesized at Georgia Tech. [5]. At Institut Teknologi Bandung (ITB) a number of control synthesis have been studied for the development of an autonomous small scale helicopter. The study includes the implementation of LQR [6], LQG [7] and algebraic approach [8]. A particular case of control for transition dynamics between hover and forward flight transition has been also designed by coefficient diagram method [9] and also by formulating the phenomenon as a hybrid system leading to switched linear control design [10].

Based on the analysis of the simulation results as well as experimental verifications [4], it has been shown that the MIMO approach is superior to the SISO design due to the fact that the coupling between variables is inherently handled. The most widely used MIMO approaches, such as LQR and $H_\infty$, however have drawbacks which make them not very amenable to practical implementation in general. These include higher than necessary order of controller, non-existence of formal parameter tuning and weight selection procedures, possible exclusion of good controllers, and difficulty in integrating state variable constraints [11].

In addition, all the above control methodologies relied on the traditional approach of solving nonlinear problem by linearizing the vehicle nonlinear dynamics over a set of operating points where a gain scheduling mechanism is then applied. The approach for the control design of an autonomous helicopter at MIT [3], for instance, used a novel gain-scheduling scheme for LQR synthesis based on discrete switching of gain tables with bumpless transfer logic to accommodate rapid changes in gain values —associated with the maneuvers— with the scheduled forward speed. The disadvantage of the gain scheduling approach in general is that the control design based on the linearized dynamics might not perform well or worse guarantee stability when it is applied beyond the vicinity of equilibrium. In contrast to this approach, the LPV control technique explicitly takes into account the change in performance due to the real-time parameter variations. The technique therefore theoretically guarantees the performance and robustness over the entire operating envelope.

In one of many robust control approaches, one can conduct numerical linearizations of the dynamics of the nonlinear helicopter model made for varying forward speed. This set of linear models is then used to construct an LFT representation of the system by element-wise interpolation of the stability and control derivatives matrices. This approach, for example, has been carried out by Riyanto [12] for the design of longitudinal flight control of the N-250 aircraft. Bendotti [13] took a similar technique for the identification of model helicopter in hover for the purpose of control design using LQG and $H_\infty$. Despite its guaranteed performance, this approach however is unnecessarily conservative since the LFT does not consider the rate of parameter variation. This shortcoming can be handled by the LPV approach however it will lead to an undesirably high computational cost [14].

In this study, we investigate a new approach implementing model identification for use in the LPV control framework developed by Bamieh [15]. The proposed framework employs scheduling procedure that exhibits similar performance of standard gain scheduling, but providing stability guarantees and constraints satisfaction between the scheduling points via robust invariant sets computation. The identification scheme employs recursive least square technique implemented on the LPV system represented by dynamics of helicopter during a transition. The airspeed as the scheduling of parameter trajectory is not assumed to vary slowly. The exclusion of slow parameter change requirement allows for the application of the algorithm for aggressive maneuvering capability without the need of expensive computation. The technique is tested numerically and will be validated in the autonomous flight of a small scale helicopter. The nonlinear model of the helicopter is simulated with X-plane software while the LPV control synthesis is conducted within the MATLAB environment. The experimental validation in the instrumented small scale helicopter is currently underway.

**DYNAMICS OF MINIATURE HELICOPTER**

The basic equations of motion for a model helicopter dynamics are derived from the Newton-Euler equations for a rigid body that has six degrees of freedom to move in space. The equations of motion with reference to body-axis are given by:

$$m\vec{V} + m(\vec{\omega} \times \vec{V}) = \vec{F} \qquad (1)$$

$$I\vec{\dot{\omega}} + (\vec{\omega} \times I\vec{\omega}) = \vec{M} \qquad (2)$$

where $\vec{V} = [u\ v\ w]^T$ is the velocities vector, $\vec{\omega} = [p\ q\ r]^T$ is the angular rates vector, $F = [X\ Y\ Z]^T$ is the vector of external forces acting on the heli c.g., $M = [L\ M\ N]^T$ is the vector of external moments, $m$ is the mass of the vehicle, and $I$ is the inertial tensor.

The above two equations can be extended to a set of translational and angular motions as follows:

$$\dot{u} = vr - wq - g\sin\theta + (X_{mr} + X_{fus})/m$$
$$\dot{v} = wp - ur + g\sin\phi\cos\theta + (Y_{mr} + Y_{fus} + Y_{tr} + Y_{vf})/m$$
$$\dot{w} = uq - vp + g\cos\phi\cos\theta + (Z_{mr} + Z_{fus} + Z_{ht})/m \qquad (3)$$
$$\dot{p} = qr(I_{yy} - I_{zz})/I_{xx} + (L_{mr} + L_{vf} + L_{tr})/I_{xx}$$
$$\dot{q} = pr(I_{zz} - I_{xx})/I_{yy} + (M_{mr} + M_{ht})/I_{yy}$$
$$\dot{r} = pq(I_{xx} - I_{yy})/I_{zz} + (-Q_e + N_{vf} + N_{tr})/I_{zz}$$

The subscript indicates the component generating the respective force or moment: main rotor ( )$_{mr}$, fuselage including fuselage aerodynamics effects ( )$_{fus}$, tail rotor ( )$_{tr}$, vertical fin ( )$_{vf}$ and horizontal tail ( )$_{ht}$.  $-Q_e$ is the torque produced by the engine to counteract the aerodynamic torque on the main rotor blades. In the above equations the cross product of inertia was neglected. It is also assumed that the fuselage center of pressure coincides with the c.g., therefore the moments created by the fuselage aerodynamic forces were neglected. See [16] for details about forces and moments expressions

To get attitude information, Euler angles can be obtained by the following kinematic relations:

$$\begin{bmatrix} \dot{\phi} \\ \dot{\theta} \\ \dot{\psi} \end{bmatrix} = \begin{bmatrix} 1 & \tan\theta\sin\phi & \tan\theta\cos\phi \\ 0 & \cos\phi & -\sin\phi \\ 0 & \sin\phi/\cos\theta & \cos\phi/\cos\theta \end{bmatrix} \begin{bmatrix} p \\ q \\ r \end{bmatrix} \quad (4)$$

For miniature helicopter, the rotor dynamics gives major influence to the vehicle rigid-body dynamics. The coupled rotor and stabilizer bar equations are lumped into one first-order equation of motion. The lateral and longitudinal flapping dynamics are given by the following first-order equations:

$$\dot{b}_1 = -p - \frac{b_1}{\tau_e} - \frac{1}{\tau_e}\frac{\partial b_1}{\partial \mu_v}\frac{v-v_w}{\Omega R} + \frac{B_{\delta lat}}{\tau_e}\delta_{lat}$$

$$\dot{a}_1 = -q - \frac{a_1}{\tau_e} + \frac{1}{\tau_e}\left(\frac{\partial a_1}{\partial \mu}\frac{u-u_w}{\Omega R} + \frac{\partial a_1}{\partial \mu_z}\frac{w-w_w}{\Omega R}\right) + \frac{A_{\delta lon}}{\tau_e}\delta_{lon}$$

(5)

where $B_{\delta lat}$ and $A_{\delta lon}$ are effective steady-state lateral and longitudinal gains from the cyclic inputs to the main rotor flap angles; $\delta_{lat}$ and $\delta_{lon}$ are the roll and pitch cyclic control inputs; $u-u_w$, $v-v_w$ and $w-w_w$ are the wind components along, respectively, X, Y and Z helicopter body axes; $\tau_e$ is the effective rotor time constant for a rotor with the stabilizer bar. Ref.[16] provides more details about Eq.(5).

Eqs. (1)-(5) represents the mathematical model for the miniature helicopter used for the simulation. To avoid singularity and improve computation efficiency the kinematic equations, Eq. (4), were mechanized using quaternions.

**IDENTIFICATION PROBLEM**

A discrete LPV system can be represented in state-space as:

$$\begin{aligned} x(k+1) &= A(p(k))x(k) + B(p(k))u(k) \\ y(k) &= C(p(k))x(k) + D(p(k))u(k) \end{aligned} \quad (6)$$

The exogenous parameter $p(k)$ can in general be measured or estimated upon operation of the system [17].

A special class of discrete-time LPV models used in this work is parameterized as follows:

$$M(\delta,p) = \frac{B(\delta,p)}{A(\delta,p)} \quad (7)$$

where $\delta$ is the delay operator and

$$\begin{aligned} B(\delta,p) &:= b_0(p) + b_1(p)\delta + \ldots + b_{n_b}(p)\delta^{n_b} \\ A(\delta,p) &:= 1 + a_1(p)\delta + \ldots + a_{n_a}(p)\delta^{n_a} \end{aligned} \quad (8)$$

$n = na + nb + 1$ is the number of parametric functions to be identified. In this case, it is assumed that the varying parameter p is a function of discrete time, i.e $p = p(k)$.
The operational notation relating input and output of the system given by Eq.(7) is expressed as

$$A(\delta,p)y(k) = B(\delta,p)u(k) \quad (9)$$

where $y$ is the output and $u$ is the input.
Our approach is to assume that the functions $\{a_i\},\{b_j\}$ are known function of the parameter $p$ of the form:

$$\begin{aligned} a_i(p) &:= \sum_{l=1}^{N} a_i^l f_l(p) \\ b_j(p) &:= \sum_{l=1}^{N} b_j^l f_l(p) \end{aligned} \quad (10)$$

where the constants $a_i^l$ and $b_j^l$ are real numbers and defined for $i = 1,\ldots, n_a$ and $j = 1,\ldots,n_b$ respectively. Therefore any particular model in this class is completely characterized by the real number $a_i^l$ and $b_j^l$. The goal of the parametric identification scheme is to find these constants from data [14]. For this general framework, many choices are possible for the functions $f_l(p)$. One particular interest is when we consider a polynomial dependence, i.e.:

$$f_l(p) = p^{l-1} \quad l = 1,\ldots,N \quad (11)$$

where the coefficient functions are polynomial of order $N-1$, i.e.

$$\begin{aligned} a_i(p) &= a_i^1 + a_i^2 p + \ldots + a_i^N p^{N-1} \\ b_j(p) &= b_j^1 + b_j^2 p + \ldots + b_j^N p^{N-1} \end{aligned} \quad (12)$$

Following the notations, assumptions and algorithms in [14], we performed the RLS identification to estimate the parameters A and B. The RLS scheme is attractive due to its suitability for real time control application.

**HELI LPV MODEL IDENTIFICATION**

To implement LPV model identification for small scale helicopter, the equation motions of the vehicle is transformed into a dynamic model as given in Eq.(9). The problem can be viewed as the identification of nonlinear systems linearized along a varying trajectory. In this case, $p$ is defined as the total velocity of the helicopter. Judging the order of the system, we select the following form of the parameters:

$$A(\delta, p) = 1 + A_1(p)\delta^{-1} + A_2(p)\delta^{-2} + A_3(p)\delta^{-3} \quad (13)$$

and for $i = 1,2,3$, $A_i(p) = a_i^1 + a_i^2 p + a_i^3 p^2$ and

$$B(\delta, p) = B_2(p)\delta^{-2} \quad (14)$$

where $B_2(\delta, p) = b_2^1 + b_2^2 p + a_2^3 p^2$

Identification is analyzed in 2nd order parameter trajectory sequences,

$$p(k) = \begin{bmatrix} 1 & \sin(\pi k/3) & \sin(\pi k/3)^2 \end{bmatrix} \quad (15)$$

The simulation was carried out with $T = 0.11408\ s$ with the range $0 \leq p \leq 13 m/s$.

**RESULTS AND DISCUSSIONS**

By using RLS algorithm with parameter trajectory sequences $p(k) = \sin(\pi k/3)$, we obtained $A$ and $B$ parameters shown in Figs. 1,3 and 5 for pitch, vertical velocity and forward velocity respectively. To validate the result of the identification, a comparison is performed between system output and the output of the identified model. Fig. 2 shows the comparison for the case of pitch. The identified model in general follows the system well enough with the exception of spikes at $t = 45$ and $54$. Figs. 4 and 6 show the comparison of output for the case of vertical and forward velocity respectively. The two figures indicate the good performance of the LPV model identification. In both cases, the output of the identified model matches the system output reasonably well.

It must be noted that at the current stage, the LPV model identification algorithm is only implemented for the SIMO case. Based on the analysis in Ref. [8], the dominant control for the longitudinal mode is pitch-cyclic. The identification scheme was then implemented for the pitch-cyclic input with the output of pitch, vertical velocity and forward velocity.

To extend the result to MIMO case, an appropriate modification in the corresponding matrix equations of the LPV model identification scheme will be necessary. Even though theoretically it is possible to generalize the algorithm for the MIMO system, the RLS implementation might be cumbersome leading to the difficulty for the practical real-time control application.

*Input pitch-cycle - output pitch* $\theta$

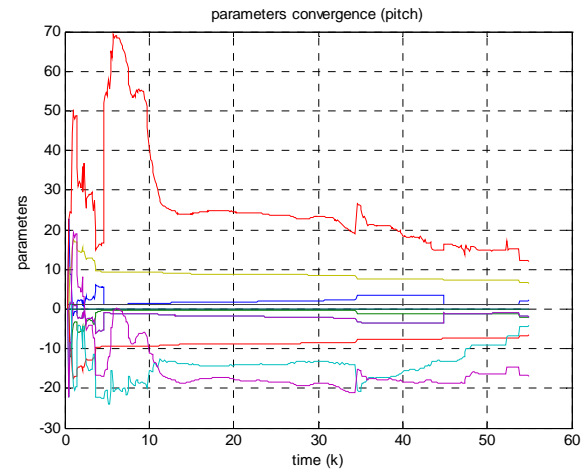

Fig.1 Parameters' convergence (pitch)

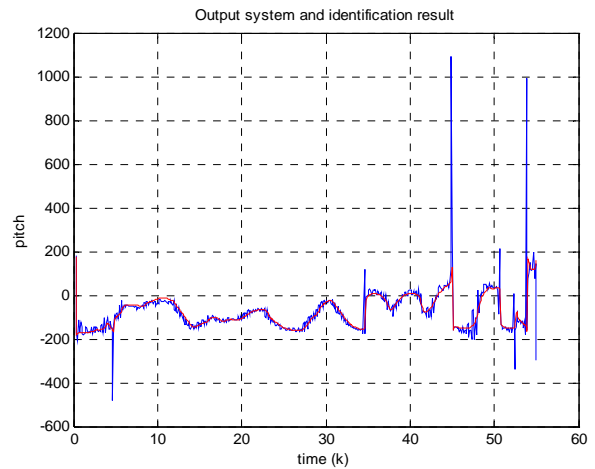

Fig.2 System output versus identification result (pitch)

Fortunately for most aerospace systems, the equations of motion typically can be decomposed into to decoupled modes each of which can be adequately stabilized or controlled by using two inputs. For helicopter, the longitudinal-vertical can be stabilized using the collective pitch and pitch cyclic. Meanwhile the lateral directional can be controlled by using the roll-cyclic and collective pedal. In the realm of LPV model identification, this fact is amount to reducing the generalization of the SIMO case only to TIMO (two input multi output) system.

*Input pitch-cycle — output velocity* w

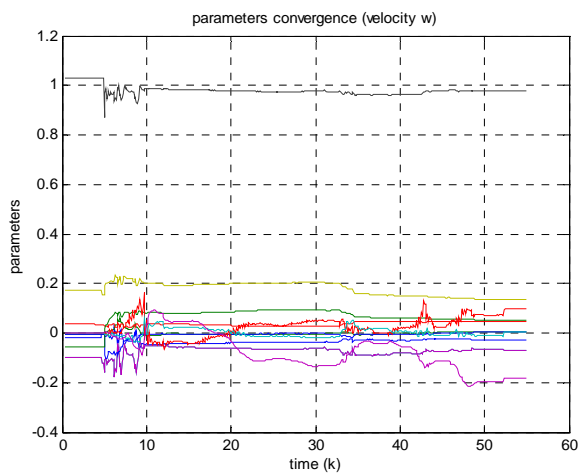

Fig.3 Parameters' convergence (vertical velocity)

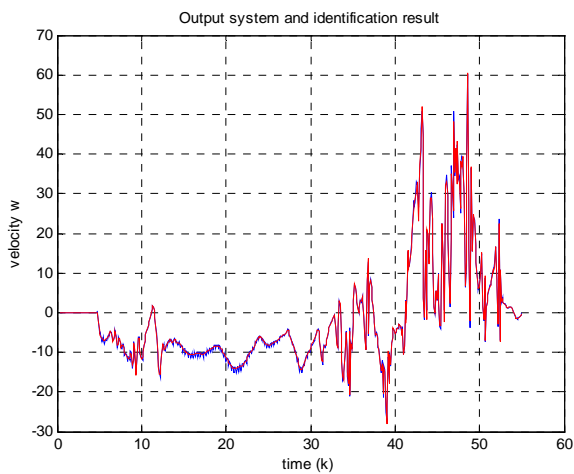

Fig.4 System output versus identification result (vertical velocity)

*Input pitch-cycle      output forward velocity* u

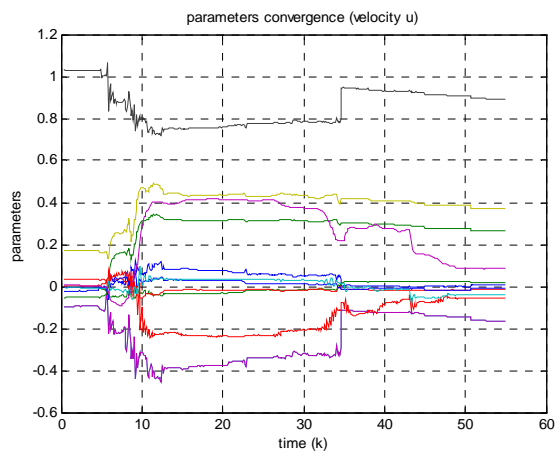

Fig.5 Parameters' convergence (forward velocity)

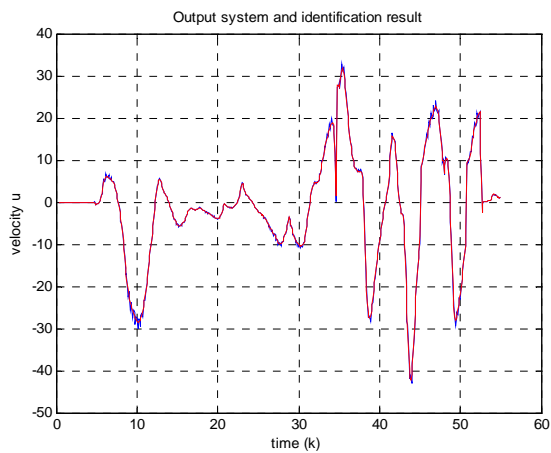

Fig.6 System output versus identification result (forward velocity)

## CONCLUSIONS

The LPV model identification scheme for a small scale helicopter control has been presented. The LPV control framework theoretically guarantees the performance and robustness over the entire operating envelope as it considers explicitly the effect of parameter variations to change of performance. In the proposed LPV model identification, there is no requirement that the scheduling parameter trajectories vary slowly while collecting enough data to identify local LTI model. This property is therefore amenable for the practical control design of small scale helicopter during aggressive maneuvers. Further work to generalize the LPV model identification framework to two-input case is necessary prior to experimental validation using an instrumented small scale helicopter.